# RARE disease detection from Capsule Endoscopic Videos based on Vision Transformers

X. Gao, C. Chien, G. Liu, A. Manullang

Department of Computer Science, Middlesex University, London, UK

Corresponding Author Email: x.gao@mdx.ac.uk

Team Name: MDXBrain

Github Repository Link: https://github.com/xiaohong1/RARE-Disease-ViT/tree/main

**Abstract**

This work is corresponding to the Gastro Competition for multi-label classification from capsule endoscopic videos (CEV). Deep learning network based on Transformers are fined-tune for this task. The based online mode is Google Vision Transformer (ViT) batch16 with 224 x 224 resolutions. In total, 17 labels are classified, which are *mouth*, *esophagus*, *stomach*, *small intestine*, *colon*, *z-line*, *pylorus*, *ileocecal valve*, *active bleeding*, *angiectasia*, *blood*, *erosion*, *erythema*, *hematin*, *lymphangioectasis*, *polyp*, and *ulcer*. For test dataset of three videos, the overall mAP @0.5 is **0.0205** whereas the overall mAP @0.95 is **0.0196**.

## 1  Motivation

Gastrointestinal tract (GI) is a 9-meter long food passage. Due to its narrow space with around 2 cm in diameter in most of the tract, diagnosis is extremely challenging if not impossible. The current state-of-the-art (SOTA) is to ask a patient to swallow a tiny capsule camera, a video capsule Endoscopy (VCE). Then VCE travels from the mouth to anus (M2A) filming the events along the tract. The video is typically of 8-hour long containing around 50,000 2D images (frames). The challenges here including not only large amount of images to inspect for clinicians but also the low quality of images. A video image has a low resolution around 512 x 512 pixels in comparison with a traditional endoscopic image of GI tract with 1900 x 2000 pixels. Significantly, the VCE images present not only artefact, e.g. blurry, bubbles, and specularity, but also food debris, which exacerbates the accuracy of diagnosis. Hence application of AI technology to assist clinical diagnosis is a key to decision making. In addition, instead of the traditional single label image classification, this task is in need of identifying anatomic regions, including *mouth*, *esophagus*, *stomach*, *small intestine*, *colon*, *z-line*, *pylorus*, and *ileocecal valve*, in addition of pathological disease findings. These pathological diseases are active *bleeding*, *angiectasia*, *blood*, *erosion*, *erythema*, *hematin*, *lymphangioectasis*, *polyp* and *ulcer*. Hence a multi-label classification model will be developed in this study.

## 2   Methods

This work corresponds to the ICPR 2026 RARE-VISION Competition, organised by Lawniczak et al. [1]. In this study, a transformer-based model will be enhanced to perform the classification. It is based on Vision Transformers (ViT) with a baseline model being ViT batch16 with a resolution of 224×224 pixels. Figure 1 depicts the overall pipeline of the work in this study.

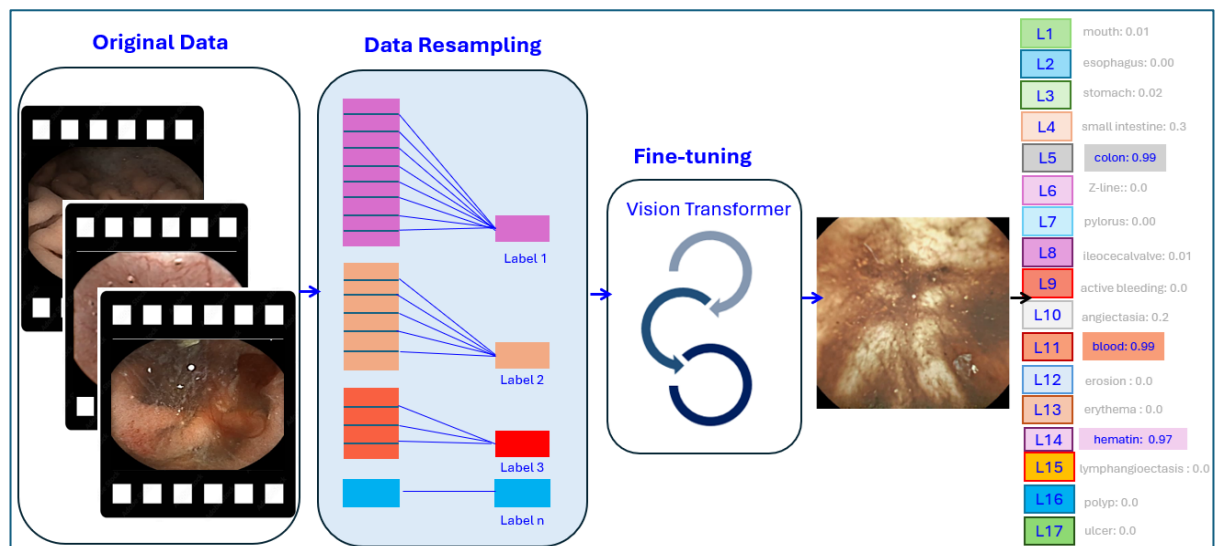

Figure 1. The overview of the model development in this study.

### 2.1   How was class imbalance handled?

The datasets are collected from the RARE Vision Competition with 17 labels including both anatomic regions and pathological findings from gastrointestinal (GI) tract. Tables 1 provides the number of of each label in this competition. It appears that the data are extremely unbalanced. For example, the data set that contains z-line only have 122 data whereas 1,878,361 data are observed with colon label. Most labels have a few thousand data. In this study, semi-supervised random under-sampling approach is applied. The priority was given to the images with multiple labels. For example, there is only one image that has the maximum label number of 5 ( 'small intestine', 'pylorus', 'activae bleeding', 'blood', 'angiectas') (Figure 2). This image was included. Otherwise, if image numbers with concerned labels were larger than required, random selection was performed. In addition, most images have only 1 label (n=2,994,127) (Table 1). In this work, all images with 4 and 5 labels were included for training. For shared label numbers being 1 to 3, a random selection was performed to ensure the data are balanced to a greater extent. As a result, around 3,000 images for each class were selected in order as illustrated in Table 1.

Table 1. Illustration of selected data for training and validation.

| Anatomy | Origainal | Training | Validation | Total Selected |
|---|---|---|---|---|
| Mouth | 2,009 | 1,697 | 402 | 2,009 |
| Esophagus | 2,256 | 1,805 | 451 | 2,256 |
| Stomach | 254,994 | 3224 | 807 | 4,031 |

| Small Intestine | 1,375,918 | 3894 | 975 | 4,869 |
|---|---|---|---|---|
| Colon | 1,878,361 | 4,347 | 1,088 | 5,435 |
| z-line | 122 | 98 | 24 | 122 |
| Pylorus | 3,183 | 2546 | 637 | 3,183 |
| Ileocecal valve | 3,692 | 2,954 | 738 | 3,692 |
| Active bleeding | 5,325 | 2,831 | 707 | ,3538 |
| Angiectasia | 16,803 | 3,683 | 921 | 4,604 |
| Blood | 391,715 | 3,556 | 889 | 4,445 |
| Erosion | 39,105 | 3,828 | 958 | 4,786 |
| Erythema | 6,228 | 3,431 | 858 | 4,289 |
| Hematin | 31,773 | 4,228 | 1,058 | 5,286 |
| Lymphangioectasis | 17,660 | 3,314 | 828 | 4,142 |
| Polyp | 18,415 | 3,241 | 810 | 4,051 |
| Ulcer | 11,428 | 3,840 | 960 | 4,800 |
| **Total** | | 42,234 | 10,890 | 53,124 |

Table 2 specifies the image numbers with shared class labels as well as the selected amount of training and validation.

In this study, the data imbalance issue was addressed based on semi-supervised under-sampling methods. For images that share more than 3 class labels, all those images were included. The proposed under-sampling approach was performed in a decision tree style, with a descending order of label numbers to examine, e.g. 5, 4, 3, 2 and 1.

Table 2. Image numbers that share the same number of class labels.

| **Multi-label** | **1** | **2** | **3** | **4** | **5** |
|---|---|---|---|---|---|
| **Image No** | 2,994,127 | 495,142 | 22,501 | 1,767 | 1 |
| **Training** | 4,236 | 12,605 | 9,838 | 1,904 | 1 |
| **Validation** | 1,016 | 3,940 | 2,141 | 517 | 0 |

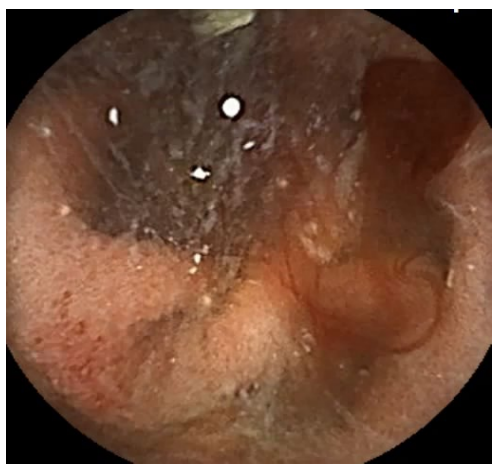

Figure 2. The image that has the largest number of labels being 5, which are *small intestine*, *pylorus*, *active bleeding*, *blood*, and *angiectasia*.

# 3  Results

The resting results are shown below with Table 3 listing the break-down details for each test video. Overall mAP @ 0.5 is 0.0205 wherelst the overall mAP @ 0.95 is 0.0196.

Table 3. Per-video mAP break down performance.

| Video ID | mAP @ 0.5 | mAP @ 0.95 |
|---|---|---|
| 001 | 0.0596 | 0.0589 |
| 002 | 0.0018 | 0.0001 |
| 003 | 0.0001 | 0 |

# 4  Summary

In summary, the team MDABrain utilized the Vision Transformer technique to perform the task of multi-label classification for RARE disease detection from VCE videos. The random under - sampling approach was employed to address dataset that were inherently imbalanced.

The team achieved an overall mAP@0.5 of 0.0205 and an overall mAP@0.95 of 0.0196.

# 5  Acknowledgments

All authors are extremly grateful to organiers of the ICPR 2026 RARE-VISION Compe-tition [1].

The authors also like to acknowledge the financial support of the British Council at the UK to allow Early Career Fellows (2024-2025) to participate this competition.
.

# References

[1] Anni Lawniczak, Manas Dhir, Maxime Le Floch, Palak Handa, and Anastasios Koulaouzidis. ICPR 2026 RARE-VISION Competition Document and Flyer. 12 2025. doi: 10.6084/m9.figshare.30884858.v3. URL https://figshare.com/articles/preprint/ICPR_2026_RARE-VISION_Competition_Document_and_Flyer/30884858 .